\documentclass[10pt,twocolumn,letterpaper]{article}

\usepackage[pagenumbers]{cvpr} 

\usepackage{graphicx}
\usepackage{amsmath}
\usepackage{amssymb}
\usepackage{booktabs}
\usepackage{multirow}

\usepackage[pagebackref,breaklinks,colorlinks]{hyperref}

\usepackage[capitalize]{cleveref}
\crefname{section}{Sec.}{Secs.}
\Crefname{section}{Section}{Sections}
\Crefname{table}{Table}{Tables}
\crefname{table}{Tab.}{Tabs.}

\def\cvprPaperID{9332} 
\def\confName{CVPR}
\def\confYear{2023}

\begin{document}

\title{Auto-Focus Contrastive Learning for Image Manipulation Detection}

\author{Wenyan Pan{$^{1}$}, Zhili Zhou{$^{2}$\thanks{Corresponding Author}}, Guangcan Liu{$^{3}$}, Teng Huang{$^{2}$}, Hongyang Yan{$^{2}$}, Q.M. Jonathan Wu{$^{4}$} \\
{$^{1}$}School of Computer and Software, Nanjing University of Information Science and Technology\\
{$^{2}$}Institute of Artificial Intelligence and Blockchain, Guangzhou University \\
{$^{3}$}School of Automation, Southeast University \\
{$^{4}$}Department of Electrical and Computer Engineering, University of Windsor \\
{\tt\small {$^{1}$}Panwy@nuist.edu.cn, {$^{2}$}zhou{\_}zhili@163.com, {$^{3}$}gcliu1982@gmail.com} \\
{\tt\small {$^{2}$}\{huangteng1220, hyang{\_}yan\}@gzhu.edu.cn, {$^{4}$}jwu@uwindsor.ca}
}
\maketitle

\begin{abstract}
   Generally, current image manipulation detection models are simply built on manipulation traces. However, we argue that those models achieve sub-optimal detection performance as it tends to: 1) distinguish the manipulation traces from a lot of noisy information within the entire image, and 2) ignore the trace relations among the pixels of each manipulated region and its surroundings. To overcome these limitations, we propose an Auto-Focus Contrastive Learning (AF-CL) network for image manipulation detection. It contains two main ideas, i.e., multi-scale view generation (MSVG) and trace relation modeling (TRM). Specifically, MSVG aims to generate a pair of views, each of which contains the manipulated region and its surroundings at a different scale, while TRM plays a role in modeling the trace relations among the pixels of each manipulated region and its surroundings for learning the discriminative representation. After learning the AF-CL network by minimizing the distance between the representations of corresponding views, the learned network is able to automatically focus on the manipulated region and its surroundings and sufficiently explore their trace relations for accurate manipulation detection. Extensive experiments demonstrate that, compared to the state-of-the-arts, AF-CL provides significant performance improvements, i.e., up to 2.5\%, 7.5\%, and 0.8\% F1 score, on CAISA, NIST, and Coverage datasets, respectively.
\end{abstract}

\begin{figure}[t]
  \centering
   \includegraphics[width=1\linewidth]{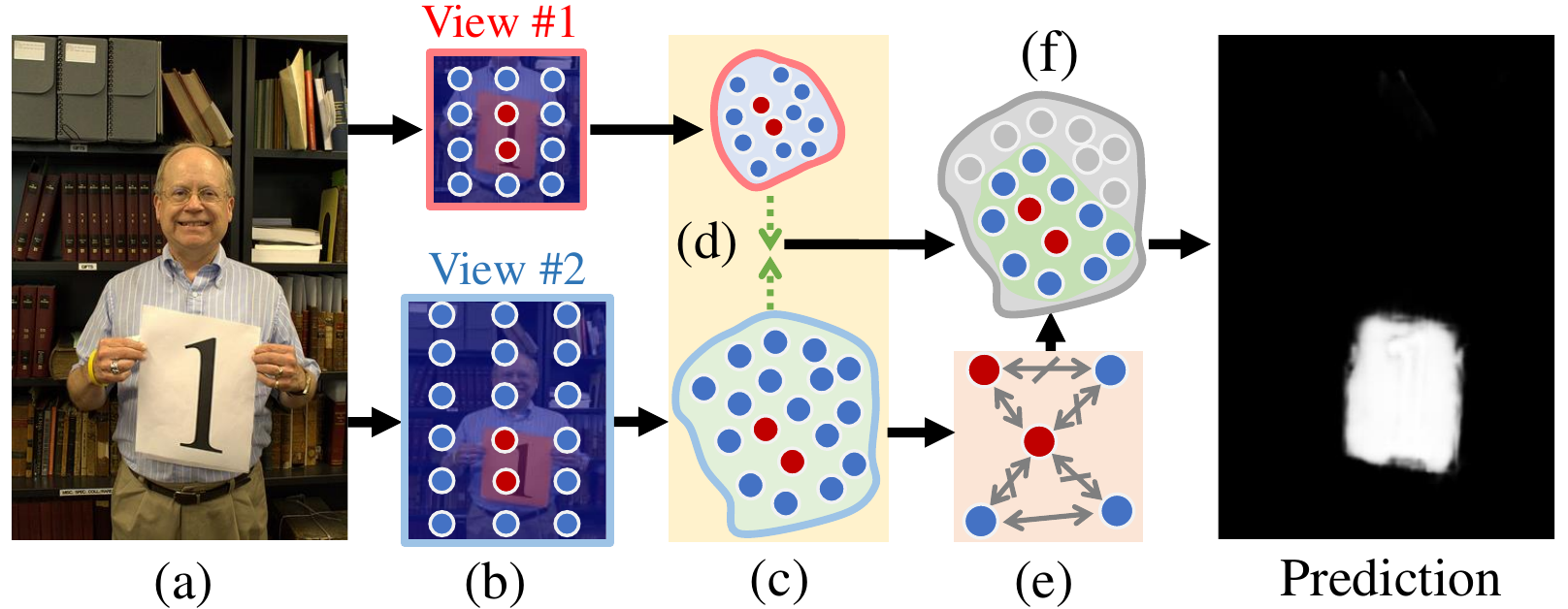}
   \caption{\textbf{Main ideas}. Current image manipulation detection models learn to map the image into an embedding space (c), yet it tends to: 1) distinguish the traces from a lot of noisy information within the entire image, and 2) ignore the trace relations among the pixels of each manipulated region and its surrounding. The manipulated and non-manipulated regions are noted by different colors. The Auto-Focus Contrastive Learning (AF-CL) network is proposed to overcome the above weaknesses, by 1) minimizing the distance ($\textcolor[RGB]{112,173,71}\dashrightarrow\textcolor[RGB]{112,173,71}\dashleftarrow$) the embedding space from two views (\textit{i.e}., (d)), and 2) modeling the trace relations by considering the similarity relations ($\textcolor[RGB]{127,127,127}\leftrightarrow$) among the same region (\textcolor{blue}{•} \textcolor{blue}{•}, \textcolor{red}{•} \textcolor{red}{•}) and dissimilarity relations ($\textcolor[RGB]{127,127,127}\nleftrightarrow$) among the different view (\textcolor{blue}{•} \textcolor{red}{•}) (\textit{i.e}., (e)). In such a way, the proposed AF-CL network is able to automatically focus on the manipulated region with its surroundings (f) and sufficiently explore their trace relations (e), thus achieving promising performance for image manipulation detection.}
   \label{Fig. 1}
\end{figure}

\section{Introduction}
\label{sec:intro}

With the increasing popularity of digital images and the rapid development of powerful image processing tools, some unscrupulous individuals can easily manipulate image content to mislead the public, resulting in serious society social issues \cite{wentao2021predicting, huh2018fighting}. To combat and reveal these illegal manipulation behaviors, it is urgently required to develop an effective image manipulation detection model to detect and locate the manipulated image regions accurately.   

Generally, current image manipulation detection methods \cite{zhou2018learning, bayar2018constrained, wu2019mantra, zhou2020generate, chen2021image, wang2022objectformer, hu2020span} employ deep neural networks to map the image into a non-linear high-dimensional embedding space to capture the manipulation traces caused by a variety of manipulation attacks, such as splicing, copy-move, and removal. However, the current models suffer from two following issues. \textbf{1)} As a lot of noisy information is usually left in the image by the blurring/filtering pre-processing with the manipulation attacks, it is still hard for those models to accurately distinguish the manipulation traces from the noisy information within the entire image. That would misguide the model to focus on irrelevant regions for manipulation detection, thus compromising the detection performance. \textbf{2)} As intrinsic characteristics of manipulated images, the trace relations among the pixels of each manipulated region and its surroundings are also beneficial for manipulation detection, but they are rarely explored. Specifically, the modification traces among the pixels within the manipulated region are more similar, while the modification traces among the pixels of the manipulated region and its surrounding are quite different. Therefore, the trace relations can describe the intrinsic characteristics of manipulated regions, and thus serve as discriminative representation to detect the manipulated regions accurately. 

\begin{figure}[t]
  \centering
   \includegraphics[width=1\linewidth]{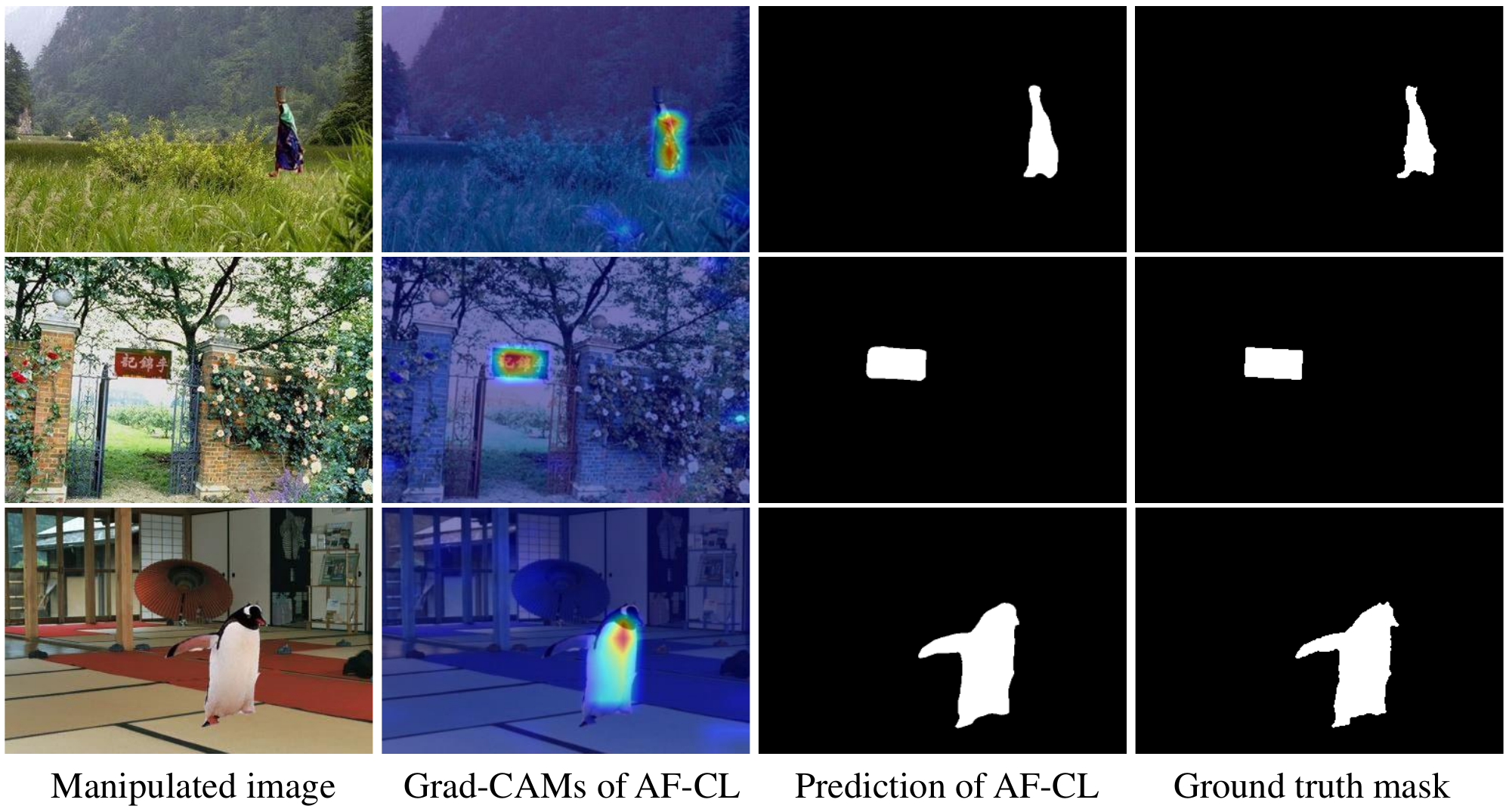}
   \caption{The toy examples of Grad-CAM \cite{selvaraju2017grad} and prediction results of the proposed AF-CL. The Grad-CAM highlights the regions on which AF-CL focuses. It is shown that AF-CL can effectively guide the model to focus on the manipulated regions and their surroundings, resulting accurate manipulation detection results.}
   \label{Fig. 2}
\end{figure}

To overcome these drawbacks of current manipulation detection models, we propose a simple-yet-effective model, \textit{i.e}., Auto-Focus Contrastive Learning (AF-CL) network, for image manipulation detection. It contains two main ideas, \textit{i.e}., multi-scale view generation (MSVG) and trace relation modeling (TRM). Specifically, the MSVG aims to generate a pair of views from a manipulated image, and each view contains the manipulated region and its surroundings at a different scale. Meanwhile, trace relation modeling (TRM) aims to model the trace relations among the pixels of the manipulated region and its surroundings across each view for producing discriminative representations. By minimizing the distance between the representations of corresponding views, the proposed AF-CL network not only focuses on the manipulated region and its surroundings, but also sufficiently explores the trace relations among the their pixels for manipulation detection. In such a way, the proposed AF-CL network can significantly improve the localization accuracy for image manipulation detection, as shown by the toy examples in Figure \ref{Fig. 2}. Our main contributions are summarized as follows:

\begin{itemize}

\item The limitations of current manipulation detection models are analyzed and revealed, including 1) distinguishing the manipulation traces from a lot of noisy information within the entire image, and 2) ignoring the manipulation trace relations among the pixels of each manipulated region and its surroundings.

\item The Auto-Focus Contrastive Learning (AF-CL) network, consisting of multi-scale view generation (MSVG) and trace relation modeling (TRM), is proposed to overcome the above limitations. The MSVG aims to generate a pair of views from a manipulated image, each of which contains the manipulated region and its surroundings at a different scale. The TRM tries to model the trace relations among the pixels of the manipulated region and its surroundings. The proposed AF-CL network can jointly utilize both components via contrastive learning to automatically focus on the manipulated region and its surroundings and explore the trace relations among their pixels for accurate manipulation detection.

\item The proposed AF-CL network is extensively evaluated with ablation studies and is compared with state-of-the-arts on widely used datasets, \textit{e.g}., CAISA, NIST, and Coverage. The results demonstrate that AF-CL outperforms the state-of-the-arts in the aspect of F1 score while maintaining good robustness to a variety of attacks without the need of pre-training on a large image manipulation dataset.

\end{itemize}

\begin{figure*}[t]
  \centering
   \includegraphics[width=0.85\linewidth]{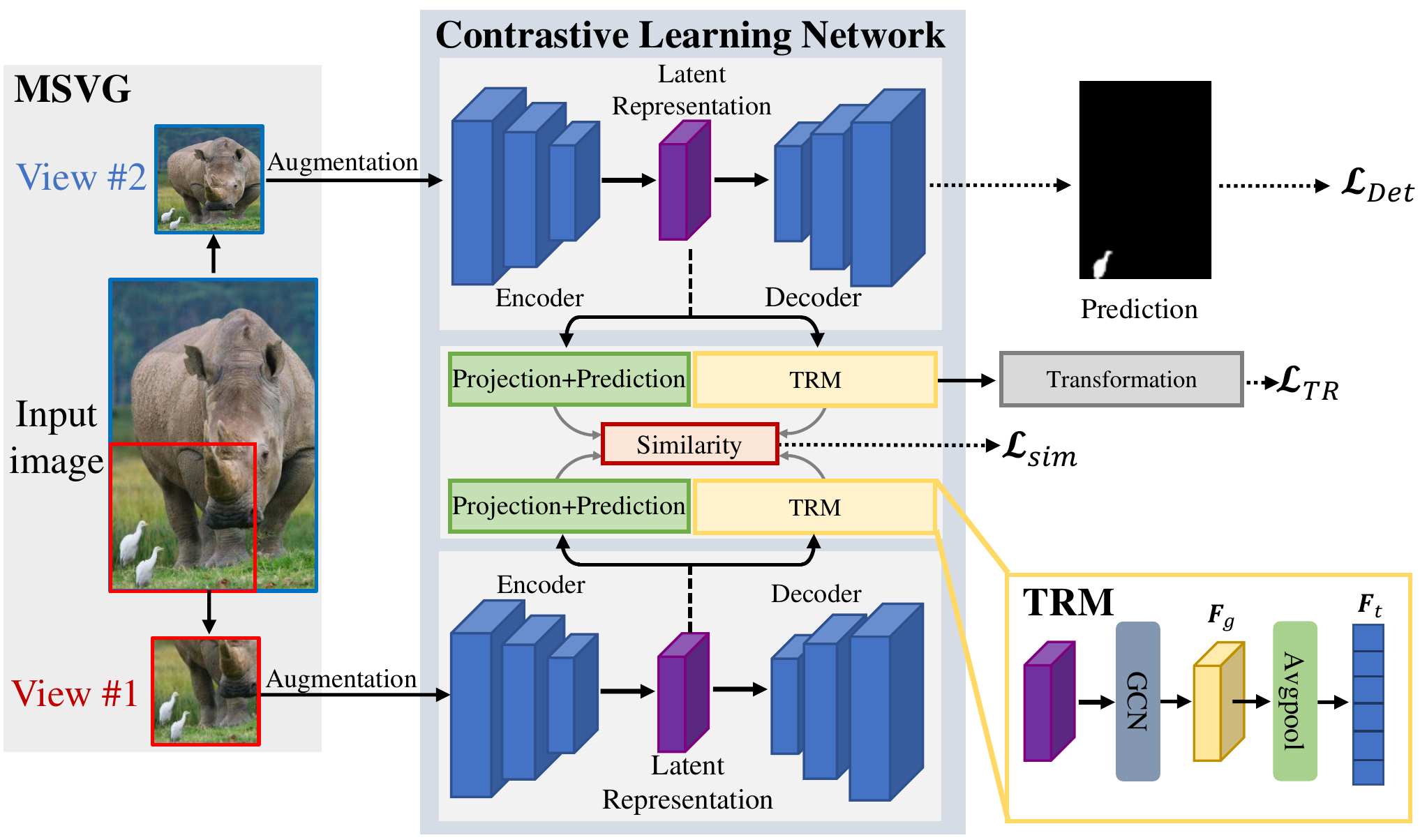}
   \caption{The overview of AF-CL. It takes a suspicious image as input, and outputs a binary prediction mask for image manipulation detection. }
   \label{Fig. 3}
\end{figure*}

\section{Related Work}
\label{Related Work}
\subsection{Image Manipulation Detection}
Image manipulation detection has been widely studied for many years, and the existing outstanding methods are mostly based on deep learning. In this context, we only focus on the deep learning method in this section.

Generally, the deep learning-based manipulation detection models are simply built on manipulation traces. Zhou \emph{et al}. \cite{zhou2018learning} designed a network architecture that contains two branches. One branch takes the RGB image as input, and another uses SRM filters \cite{fridrich2012rich} to filter the image content to extract the manipulation trace features. Bayar \emph{et al}. \cite{bayar2018constrained} designed a trainable high-pass filter-like layer, \textit{i.e}., constrained convolutional layer, to suppress the image content and then learn the manipulation feature adaptively. Wu \emph{et al}. \cite{wu2019mantra} and Hu \emph{et al}. \cite{hu2020span} adopted a constrained convolutional layer and SRM filters to learn the trace features from a different view and fuse them for further detection. Li \emph{et al}. \cite{li2019localization} proposed a two-branch network, which combines Convolutional Block Attention Module (CBAM) \cite{woo2018cbam} and SRM to discover manipulation traces. Zhou \emph{et al}. \cite{zhou2020generate} designed an edge detection and refinement branch to capture the boundary traces around the manipulated region. Chen \emph{et al}. \cite{chen2021image} designed an edge-supervised branch (ESB) to exploit boundary traces around the manipulated region. Liu \emph{et al}. \cite{liu2022pscc} designed a progressive spatio-channel correlation network (PSCC-Net) to locate the manipulated region in a coarse-to-fine fashion. Some other models transform the manipulation image into the frequency domain to capture the manipulation traces. Wang \emph{et al}. \cite{wang2022objectformer} proposed a network architecture that jointly captures manipulation traces from both the RGB domain and the frequency domain. However, as mentioned above, these models generally suffer from the following problems: 1) a lot of noisy information within the entire image makes it hard to focus on the manipulated region for accurate manipulation detection. and 2) The trace relations among the pixels of each manipulated region and its surroundings are ignored, which are very important for manipulation detection. Therefore, this paper proposes the AF-CL network to overcome the above limitations for image manipulation detection.

\subsection{Contrastive Learning}

Contrastive learning has achieved remarkable performances in the computer vision community \cite{chen2020simple, chen2020improved, chen2021exploring, grill2020bootstrap, he2020momentum}. The previous contrastive learning networks, such as SimCLR \cite{chen2020simple} and MoCo \cite{chen2020improved}, require negative sample pairs, large batches, and a momentum encoder. Besides these networks, BYOL \cite{grill2020bootstrap} relies only on positive pairs and can achieve desirable performance. The SimSiam \cite{chen2021exploring} further designed a network based on the Siamese structure to learn image representation without negative sample pairs, large batches, and a momentum encoder.

Some methods have demonstrated that the proper image views could guide the model to focus on the target objects. Zhao \emph{et al}. \cite{zhao2021distilling} designed an augmentation strategy by copying-and-pasting the foreground onto various backgrounds to guide the model to focus on the foreground. Other methods extended the self-supervised contrastive learning to the supervised setting \cite{khosla2020supervised} to leverage label information effectively. Inspired by these methods, we propose the AF-CL network, which is able to automatically focus on manipulated regions to facilitate image manipulation detection.

\begin{figure*}[t]
  \centering
   \includegraphics[width=1\linewidth]{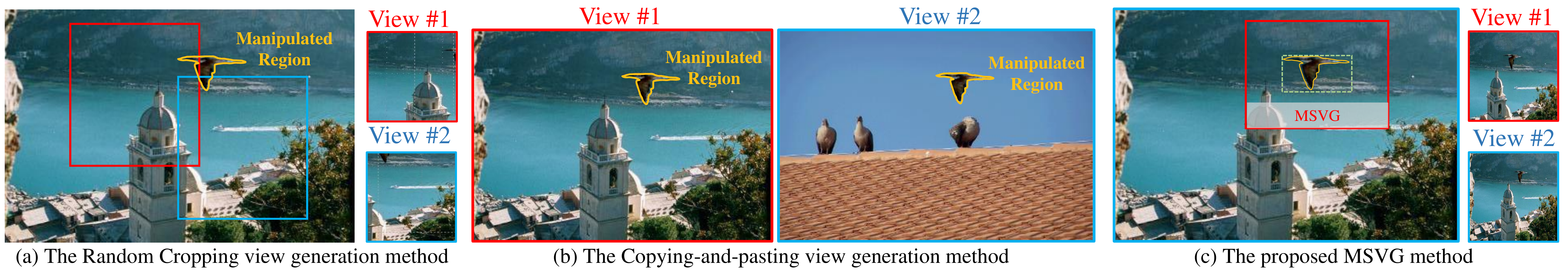}
   \caption{The different view generation methods. (a) The random cropping generates a pair of views, but it is prone to lose a part of manipulated region. (b) The copying-and-pasting generates a pair of views by copying the same manipulated region and pasting it into different backgrounds. Minimizing the distance between the corresponding views could guide the model to disregard the non-manipulated region and focus on the manipulated region, but ignores the trace relations among pixels of the manipulated and its surroundings. (c) The proposed MSVG generates a pair of views at different scales, each of which includes the manipulated region and its surroundings.}
   \label{Fig. 4}
\end{figure*}

\begin{figure}[t]
  \centering
   \includegraphics[width=0.80\linewidth]{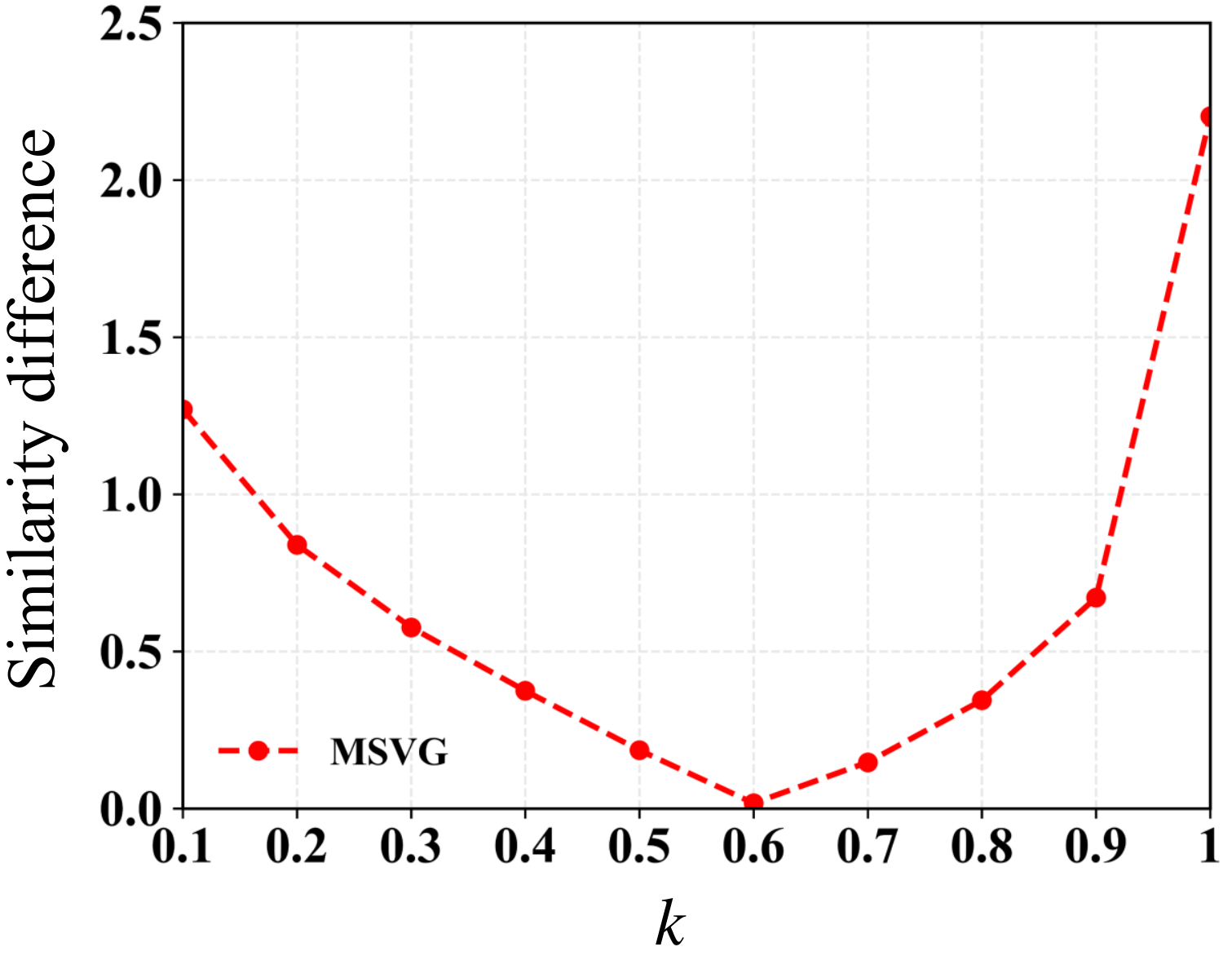}
   \caption{The change of the similarity difference and with different values of $k$.}
   \label{Fig. 5}
\end{figure}

\section{The Proposed AF-CL Network}

\subsection{Overall Framework}

The proposed AF-CL network consists of two parts: 1) The multi-scale view generation (MSVG) method, which aims to generate a pair of views containing the manipulated region and its surroundings at a different scale, and 2) The contrastive learning network, including the encoder-decoder network, projection and prediction head, and trace relation modeling (TRM). The encoder-decoder network is used to learn the latent representations and decodes the representations for final prediction. The projection and prediction head transform the representations from a pair of views to two positive representation pairs $F_d$, $F_d^\prime$, $F_z$, $F_z^\prime$, respectively. The trace relation modeling (TRM) is used to learn the trace relations among the pixels of the manipulated region and its surroundings, which denoted as $F_t$ and $F_t^\prime$, respectively. We minimize the distance of transformed representations and trace relations between the two generated views, and also use the representations to locate the manipulated region. The overall framework of AF-CL is given in Figure \ref{Fig. 3}. 

\subsection{Multi-Scale View Generation}
\label{subsec:MSVG}

The essential step of the proposed AF-CL is to generate a pair of views from each given image. There are some off-the-shelf view generation methods, \textit{e.g}., random cropping and copying-and-pasting \cite{zhao2021distilling}. However, they either lose a part of the manipulated region or ignore trace relations among pixels of the manipulated and its surroundings (Figure \ref{Fig. 4}(a) and \ref{Fig. 4}(b)). To avoid the above issues, we design the MSVG method to generate multi-scale views, including a large-scale view, which is the entire manipulated image (view$\#$2 in Figure \ref{Fig. 4}(c)), and a small-scale view, which is centered at the manipulated region (view$\#$1 in Figure \ref{Fig. 4}(c)) and is represented by:

\begin{equation}
\label{eq.1}
    (x_l, y_l, x_r, y_r) = \mathbb{R}_{\textit{MSVG}}(I,GT,k)
\end{equation}

\noindent where $\mathbb{R}_{\textit{MSVG}}(\cdot,\cdot,\cdot)$ is the view generation function that returns the top-left and bottom-right corners of the generated view, $I$ and $GT$ denote the manipulated image and the ground-truth mask of the manipulated region with a size of $h \times w$, respectively. $k$ represents the size ratio of the small-scale view (view$\#$2) to the entire image, and $k\in\left[0.1,0.2,...1\right]$.

The scale of view$\#$2 is controlled by $k$: a larger $k$ leads to a larger view$\#$2, but increases the possibility of introducing more irrelevant information; a smaller $k$ leads to a smaller view$\#$2, which contains a less non-manipulated region (\textit{e.g}., MSVG generates the view with minimum scale, see the green box in Figure \ref{Fig. 4}(c)). To select a proper $k$ for our task, we compute the similarity between view$\#$1 and view$\#$2 (denoted as $S_{\#1\rightarrow\#2}$) as well as the similarity between view$\#$2 and minimum view$\#$2 (denoted as $S_{\#2\rightarrow min\#2}$) of many manipulated images by the $l_2$ distance via a standard ResNet-50. If the difference between the two computed similarities is equal to a minimum value, we assume that a good trade-off between sizes of the non-manipulated region and manipulated region is obtained, and thus the corresponding proper $k$ is determined. We set $k$ as different values and test the similarity difference $(\left|S_{\#1\rightarrow\#2}-S_{\#2\rightarrow min\#2}\right|)$, and the result is shown in Figure \ref{Fig. 5}. It can be clearly observed that the similarity difference reaches a minimum value when $k=0.6$, and thus we set $k=0.6$ as the proper value in the proposed MSVG method. To further prove this assumption, we investigate the impact of $k$ in Section \ref{subsec:Ablation Studies}.

\subsection{Trace Relation Modeling}

In this paper, we implement the trace relation modeling by graph convolutional network (GCN) \cite{welling2016semi} to explore the trace relations among the pixels of the manipulated region and its surroundings, as GCN can model the relationships between objects \cite{zhao2019t, zhao2021graphfpn, zhang2019dual}.

We take a latent representation $L \in \mathbb{R}^{B \times C \times H \times W}$ learned from the view as input, where $B$, $C$, $W$, and $H$ denotes the batch size, channel, width, and height of the latent representation, respectively. We first transform the $L$ to a new representation $L' \in \mathbb{R}^{B \times C \times H/2 \times W/2}$ in the new space by down-sampling. Then, we use a fully connected graph $G = (V, E)$ to describe the trace relations among pixels of the manipulated and its surroundings, and treat each grid in the latent representation as a node, where $V$ is a set of nodes, $V=\left\{ v_1, v_2, ..., v_n \right\}$, $N \in \mathbb{R}^{(H \times W) / 4}$ is the number of node, and $E$ is a set of edges. The adjacency matrix $\textbf{\textit{A}} \in \mathbb{R}^{N \times N}$ is used to represent the connection between all nodes, except for the nodes with themselves. Thus, the adjacency matrix is a (0,1)-matrix with zeros on its main diagonal. The element is 0 if there is no connection between nodes, and one denotes there is a connection. Thus, modeling the trace relations $F_g$ can be considered as learning the mapping function $f$ on the graph $G$ and transformed latent representation $L'$ with a 2-layer GCN model which can be expressed as:

\begin{equation}
\label{eq.3}
  F_g = f(L', \textbf{\textit{A}})=\mathcal{T}\left(\sigma_s\left( \widehat{\textbf{\emph{A}}} \sigma_r \left( \widehat{\textbf{\emph{A}}} L' \textbf{\emph{M}}_0 \right )\textbf{\emph{M}}_1\right)\right)
\end{equation}

\begin{equation}
\label{eq.4}
    \widehat{\textbf{\emph{A}}}= \widetilde{\textbf{\emph{D}}}^{-1/2} \widetilde{\textbf{\emph{A}}} \widetilde{\textbf{\emph{D}}}^{-1/2}
\end{equation}

\noindent where $\textbf{\emph{A}}$ denotes the adjacency matrix. $\textbf{\emph{M}}_0$ and $\textbf{\emph{M}}_1$ denote the weight matrix for GCN in the first and second layer. $\sigma_s$ and $\sigma_r$ denote the softmax and ReLU activation function, respectively. $ \widehat{\textbf{\emph{A}}}$ denotes the prepossessing step, $ \widetilde{\textbf{\emph{A}}}=\textbf{\emph{A}}+\textbf{\emph{I}}_N$ denotes an adjacency matrix with self-loops, and $\textbf{\emph{I}}_N$ denotes an identity matrix. $\widetilde{\textbf{\emph{D}}}$ denotes is the diagonal node degree matrix of $\widetilde{\textbf{\emph{A}}}$. $\mathcal{T}$ denotes the reshaping operation that transforms the trace relations with the dimension of $C \times B \times N$ into the $B \times C \times H \times W$. 

Finally, the learned trace relations from two views denoting as $F_g$ and $F'_g$, we transform them by average pooling layer, denoted as $Avg$. The transformed trace relations denoted as $F_t=Avg(F_g)$ and $F'_t=Avg(F'_g)$. We minimize the distance between these trace relations by:

\begin{equation}
\label{eq.5}
    \mathcal{L}_{simg}(F_t, F'_t)=\left\|F_t-F'_t\right\|_{1}
\end{equation}

\noindent where $\left\| \cdot \right\|_{1}$ is $\ell_1$-norm. In this way, the model will pay more attention to the manipulated region and its surroundings, and sufficiently explore the trace relations among the pixels of the manipulated region and its surroundings to learn discriminative features.

\subsection{Loss function}

The total loss function includes three different losses: the similarity loss to minimize the distance between generated views, the trace relation loss to supervise the trace relations modeling, and the detection loss to guarantee the prediction performance.

{\bf Similarity loss}. Our main goal is to encourage the transformed latent representation pairs $F_d$, $F_d^\prime$, $F_z$, $F_z^\prime$ and the modeled trace relations pairs $F_t$, $F_t^\prime$ from different views to be similar for guiding the model focus on the manipulated region. To this end, the similarity loss $\mathcal{L}_{sim}$ is defined as follows, 

\begin{equation}
\label{eq.6}
    \mathcal{L}_{simr}=-\frac{1}{2}\left(\frac{F_{d}}{\left\|F_{d}\right\|_{2}} \cdot \frac{F_{z}^{\prime}}{\left\|F_{z}^{\prime}\right\|_{2}}+\frac{F_{d}^{\prime}}{\left\|F_{d}^{\prime}\right\|_{2}} \cdot \frac{F_{z}}{\left\|F_{z}\right\|_{2}}\right)
\end{equation}

\begin{equation}
\label{eq.7}
    \mathcal{L}_{sim}= \sum_{\omega=0}^{\Omega}\mathcal{L}_{simr} + \sum_{\omega=0}^{\Omega}\mathcal{L}_{simg}
\end{equation}

\noindent where $\Omega$ is the number of training samples.

{\bf Trace relations loss}. In the training process, we perform trace relation loss on trace relation $F_t$ and the ground truth masks $GT$ to supervise the modeled trace relations. We employ the up-sampling to transform the $F_t$ into $H_t$ to match the $GT$. In this way, the trace relations can be guaranteed to facilitate image manipulation detection, which can be formulated as

\begin{equation}
\label{eq.8}
    \mathcal{L}_{TR}=-\sum_{\omega=0}^{\Omega}(GT \log (H_t)+(1-GT) \log (1-H_t)
\end{equation}

{\bf Detection loss}. For locating the manipulated region in pixel-level, we adopt the hybrid loss $\mathcal{L}_{Det}$ \cite{zhou2019unet++} to supervise the final prediction, which consists of pixel-wise cross-entropy loss and soft dice-coefficient loss.

\begin{equation}
\resizebox{.9\linewidth}{!}{$
    \displaystyle
    \mathcal{L}_{Det}(GT, P)=-\frac{1}{\Theta} \sum_{\gamma=1}^{\Gamma} \sum_{\theta=1}^{\Theta}\left(gt_{\theta, \gamma} \log p_{\theta, \gamma}+\frac{2 gt_{\theta, \gamma} p_{\theta, \gamma}}{gt_{\theta, \gamma}^2+p_{\theta, \gamma}^2}\right)
$}
\end{equation}

\noindent where $gt_{\theta, \gamma} \in GT$ and $p_{\theta, \gamma} \in P$ denote the ground-truth masks and predicted probabilities for class $\gamma$ and $\theta^{th}$ pixel in the batch, respectively, and $\Theta$ is the number of pixels in one batch.

{\bf Total loss function}. The total loss function $\mathcal{L}_{total}$ is a weighted sum of similarity loss $\mathcal{L}_{simr}$, trace relations loss $\mathcal{L}_{TR}$ and detection loss $\mathcal{L}_{Det}$, as follows,

\begin{equation}
\label{eq.8}
    \mathcal{L}_{total}= \mathcal{L}_{simr} + \lambda_{TR} \mathcal{L}_{TR} + \lambda_{Det} \mathcal{L}_{Det}
\end{equation}

\noindent where, $\lambda_{TR}$ and $\lambda_{Det}$ are weights for balancing different loss terms. In the training process, $\lambda_{TR}$ and $\lambda_{Det}$ is set to 0.5. 

\section{Experiments}

\subsection{Datasets \& Evaluation metrics}
We evaluate the proposed method on three widely used datasets, \textit{i.e}., CASIA \cite{dong2013casia}, NIST16 \cite{guan2019mfc}, and Coverage \cite{wen2016coverage}. The CASIA consists of CASIA 1.0 and CASIA 2.0, which contain 921 and 5124 manipulated images for testing and training, respectively. The NIST16 dataset involves copy-move, splicing, and removal manipulations, including 404 training and 160 testing images. The Coverage dataset contains 100 manipulated images, including 75 training and 25 testing images.  

To evaluate the performance of the proposed AF-CL, following the \cite{hu2020span, wang2022objectformer, liu2022pscc}, we report the image-level F1 score and Area Under Curve (AUC) as the evaluation metric. It is worth noting that, different from the previous method of binarizing the prediction mask, which uses an equal error rate (EER) threshold \cite{wang2022objectformer, liu2022pscc} based on the receiver operating characteristic curve (ROC), we directly set the prediction threshold to 0.5 to obtain a binary prediction mask. The binary prediction masks are compared with ground-truth masks to compute the F1 score, while the prediction masks without binarization are used to calculate AUC.

\begin{figure}[t]
  \centering
   \includegraphics[width=0.85\linewidth]{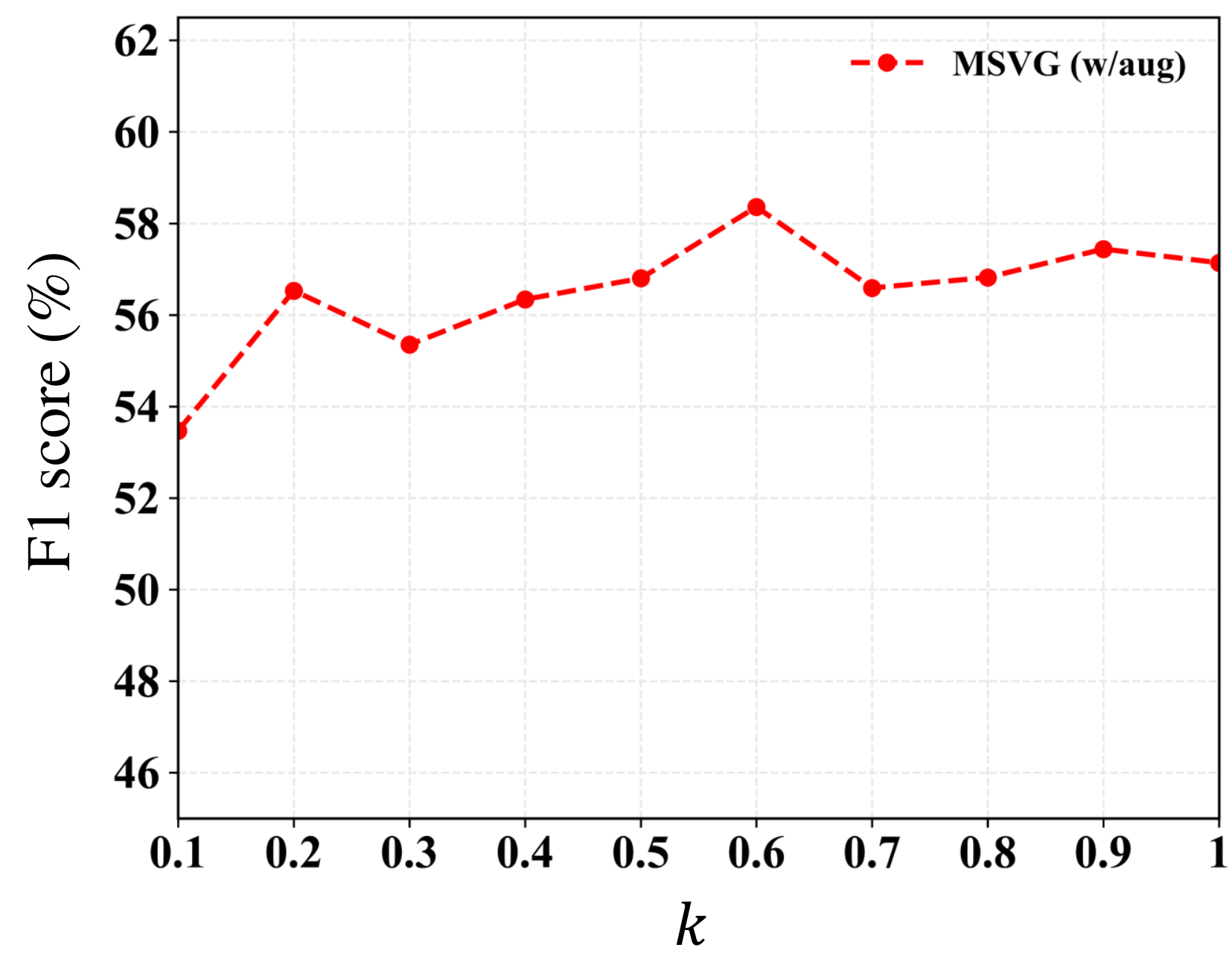}
   \caption{The impact of $k$ on the MSVG in the aspect of F1 score. Here, we adopt data augmentation strategies, denoted as w/aug.}
   \label{Fig. 6}
\end{figure}

\subsection{Implementation Details \& Baseline} 

The AF-CL is implemented on the Pytorch with an NVIDIA RTX 3090 GPU. We adopt the SimSiam \cite{chen2021exploring} contrastive learning frameworks for the network architecture and extend it to a supervised fully-supervised setting. At the training stage, the input size of the image is set to $512 \times 512$, and training is terminated after 100 epochs for the CASIA dataset, 300 epochs for the NIST16 and Coverage dataset with a batch size of 10. We use the SGD optimizer with a momentum of 0.9 and weight decay of 0.0001. Following the \cite{chen2021exploring}, we set the learning rate of $ lr \times$ batch size / 256, with a base $lr=0.05$. Following previous contrastive learning method \cite{chen2020simple, chen2020improved, chen2021exploring, grill2020bootstrap, he2020momentum}, we choose the augmentations including Flip, ColorJitter, Rotate, Grayscale, and Gauss Noise. We choose the ResNet152 network with the Feature Pyramid Network \cite{lin2017feature} as encoder-decoder and use the pre-trained model on ImageNet \cite{deng2009imagenet} to initialize the encoder-decoder to improve the training stability.

Note that the current image manipulation detection models are required to be pre-trained on a large synthesized image manipulation dataset. Instead, we directly train the proposed AF-CL on the training set and evaluate the test set of these datasets. We compare the proposed AF-CL network with state-of-the-art models, including RGB-N \cite{zhou2018learning}, SPAN\cite{hu2020span}, MantraNet\cite{wu2019mantra}, PSCCNet \cite{liu2022pscc}, and ObjectFormer \cite{wang2022objectformer}. Those models are simply built on the manipulation trace across the entire image to locate the manipulated region.

\begin{table}
  \centering
  \begin{tabular}{@{}lc@{}}
    \toprule
    View augmentation method & \multicolumn{1}{c}{F1 score} \\
    \midrule
    Copying-and-pasting & \multicolumn{1}{c}{17.25\%} \\
    RandomCrop & \multicolumn{1}{c}{48.32\%}\\
    MSVG (min) & \multicolumn{1}{c}{52.28\%}\\
    MSVG ($k$=0.6) & \multicolumn{1}{c}{\textbf{58.36}\%}\\
    \bottomrule
  \end{tabular}
  \caption{The F1 score of different view generation methods for image manipulation detection. MSVG (min) is the proposed MSVG method that generates the view with minimum scale, see the green box in Figure \ref{Fig. 4}(c)).}
  \label{Table 1}
\end{table}

\begin{table}
  \centering
  \begin{tabular}{@{}ccc@{}}
    \toprule
     MSVG & TRM & F1 score ($k$=0.6) \\ \midrule
     \checkmark &     &        58.36\%      \\
     \checkmark&    \checkmark &     \textbf{60.43}\%        \\ \bottomrule
  \end{tabular}
  \caption{Ablation studies on CASIA using different variants of AF-CL. Here, we adopt data augmentation strategies, denoted as w/aug. The F1 score is reported.}
  \label{Table 2}
\end{table}


\begin{table*}
\setlength\tabcolsep{8pt}
  \centering
  \scalebox{1}{
  \begin{tabular}{lccccccc}
    \toprule
    \multirow{2}{*}{Method} & \multicolumn{1}{c}{Pre-train} & \multicolumn{2}{c}{CASIA} & \multicolumn{2}{c}{NIST16} & 
    \multicolumn{2}{c}{Coverage} \\
    \multicolumn{1}{c}{} & \multicolumn{1}{c}{Data} & \multicolumn{1}{c}{F1} & \multicolumn{1}{c}{AUC} & \multicolumn{1}{c}{F1} & \multicolumn{1}{c}{AUC} & 
    \multicolumn{1}{c}{F1} & 
    \multicolumn{1}{c}{AUC} \\ \midrule
        RGB-N \cite{zhou2018learning} & 42K & 40.8 & 79.5 & 72.2 & 93.7 & 43.7 & 81.7 \\
        SPAN \cite{hu2020span} & 96K & 38.2 & 83.8 & 58.2 & 96.1 & 55.8 & 93.7 \\
        PSCCNet \cite{liu2022pscc} & 100K & 55.4 & 87.5 & 81.9 & 99.6 & 72.3 & 94.1 \\
        ObjectFormer \cite{wang2022objectformer} & 62K & 57.9 & 88.2 & 82.4 & 99.6 & 75.8 & 95.7\\
        AF-CL (Ours) & \textbf{0K} & \textbf{60.4} & \textbf{90.2} & \textbf{89.9} & 99.5 & 45.9 & 89.1\\
        AF-CL (Ours) & \textbf{5K} & - & - & - & - & \textbf{76.6} & \textbf{95.8}\\
    \bottomrule
  \end{tabular}
  }
  \caption{The performance comparison between the proposed AF-CL and the state-of-the-arts on CASIA, NIST16, and Coverage datasets. The F1 score and AUC are reported (in \%).}
  \label{Table 3}
\end{table*}

\subsection{Ablation Studies}
\label{subsec:Ablation Studies}

This section introduces the ablation experiments to validate the effectiveness of each component in our method, \textit{i.e}., multi-scale view generation and trace relation modeling. We conduct the experiments on CASIA 2.0, and report the F1 score on CASIA 1.0 dataset. 
\\ \hspace*{\fill} \\
\noindent {\bf Multi-Scale View Generation.} To verify the effectiveness of the proposed MSVG method for image manipulation detection, we conduct an experiment to compare the MSVG with other view generation methods. We report the F1 score in CASIA 1.0 in Table \ref{Table 1}. We have two observations: 1) the performance of the MSVG with $k=0.6$ is better than that minimum scale generated by the MSVG, and the MSVG performs better than the random cropping and copying-and-pasting generation method; 2) the copying-and-pasting generation method fails to locate the manipulated region, since it only takes the manipulated region into account. These observations demonstrate the effectiveness of the proposed MSVG, and the generated view containing manipulated region and its surrounding is essential for AF-CL.

Moreover, we study the impact of $k$ on the F1 score. Here, we adopt the augmentation strategies, including Flip, ColorJitter, Rotate, Grayscale, and Gauss Noise. Experimental results are reported in Figure \ref{Fig. 6}. It can be observed that the performance first increases with the increase of $k$. As $k$ is higher than 0.6, the performance starts to fall. The result demonstrates our assumption in section \ref{subsec:MSVG}. We choose $k=0.6$ as the default setting in the following experiments.
\\ \hspace*{\fill} \\
\noindent {\bf Trace Relation Modeling.}
This section investigates the impact of the proposed trace relation modeling (TRM). Results are shown in Table \ref{Table 2}. We observe that the proposed AF-CL (MSVG+TRM) achieves 60.43\% in F1 score on CASIA 1.0 dataset. This demonstrates that trace relations are important in learning discriminative features for image manipulation detection.

\subsection{Comparison with State-of-the-art}

In this section, we compare the proposed AF-CL performance with the state-of-the-arts on CAISA, NIST16, and Coverage. As shown in Table \ref{Table 3}, even if the proposed AF-CL does not pre-train on the large dataset, it achieves more promising performance compared with the state-of-the-arts. For the CASIA dataset, AF-CL outperforms by 2.5\% F1 score and 2.0\% AUC transformer-based method ObjectFormer \cite{wang2022objectformer}. On the NIST16 dataset, AF-CL attains a state-of-the-art performance of 89.9\% in F1 score, outperforming by a margin of 7.5\%. Despite not pre-training on a large synthesized dataset, AF-CL achieves comparable AUC performance compared to the state-of-the-arts.

The Coverage dataset only contains 100 manipulation images, making it challenging to train this dataset without pre-train. To address this problem, we use the annotations provided in Coverage dataset to copy and paste an object within the same image to generate 144 additional training images. Then we adopted the best model in CASIA dataset as a pre-trained model to finetune the training set of the Coverage dataset and evaluate its test set. As shown in Table \ref{Table 3}, the proposed AF-CL achieves 45.9\% and 89.1\% in F1 score and AUC. The performance of AF-CL increases to 76.6\% and 95.8\% in the F1 score and AUC by finetuning on the pre-trained model. Despite using the pre-trained model on a small dataset, AF-CL outperforms ObjectFormer by 0.8\% F1 score and 0.1\% AUC.

\subsection{Robustness Evaluation}

\begin{table}
  \centering
  \scalebox{0.95}{
  \begin{tabular}{lccc}
    \toprule
    Distortion & ObjectFormer & AF-CL(Ours) \\ \midrule
    w/o distortion & 87.18 & 99.53 \\ \midrule 
   Resize(0.78 $\times$) & 87.17 $\downarrow \textcolor{red}{0.01}$ & 99.54 $\uparrow \textcolor{blue}{0.01}$ \\
   Resize(0.25 $\times$) & 86.33 $\downarrow\textcolor{red}{0.85}$ & 99.46 $\downarrow \textcolor{red}{\textbf{0.07}}$\\
   GaussianBlur(\textit{kernel}=3) & 85.97 $\downarrow \textcolor{red}{1.21}$ & 99.53 $\downarrow \textcolor{red}{\textbf{0.00}}$\\
   GaussionBlur(\textit{kernel}=15) & 80.26 $\downarrow \textcolor{red}{6.92}$ & 99.23 $\downarrow \textcolor{red}{\textbf{0.30}}$\\
   GaussionNoise($\sigma=3$){*} & 79.58 $\downarrow \textcolor{red}{7.60}$ & 99.50 $\downarrow \textcolor{red}{\textbf{0.03}}$\\
   GaussionNoise($\sigma=15$){*} & 78.15 $\downarrow \textcolor{red}{9.03}$ & 99.30 $\downarrow \textcolor{red}{\textbf{0.20}}$\\
   JPEG($q$=50) & 86.24 $\downarrow \textcolor{red}{0.94}$ & 99.50 $\downarrow \textcolor{red}{\textbf{0.03}}$\\
   JPEG($q$=100) & 86.37 $\downarrow \textcolor{red}{0.94}$ & 99.52 $\downarrow \textcolor{red}{\textbf{0.01}}$ \\
    \bottomrule
  \end{tabular}}
  \caption{The robustness comparison between the proposed AF-CL and the ObjectFormer \cite{wang2022objectformer} on NIST16 dataset. The F1 score and AUC are reported (in \%). {*}: The Gaussian noise is adopted for the image augmentation in the training process.}
  \label{Table 4}
\end{table}

\begin{figure*}[t]
  \centering
   \includegraphics[width=0.7\linewidth]{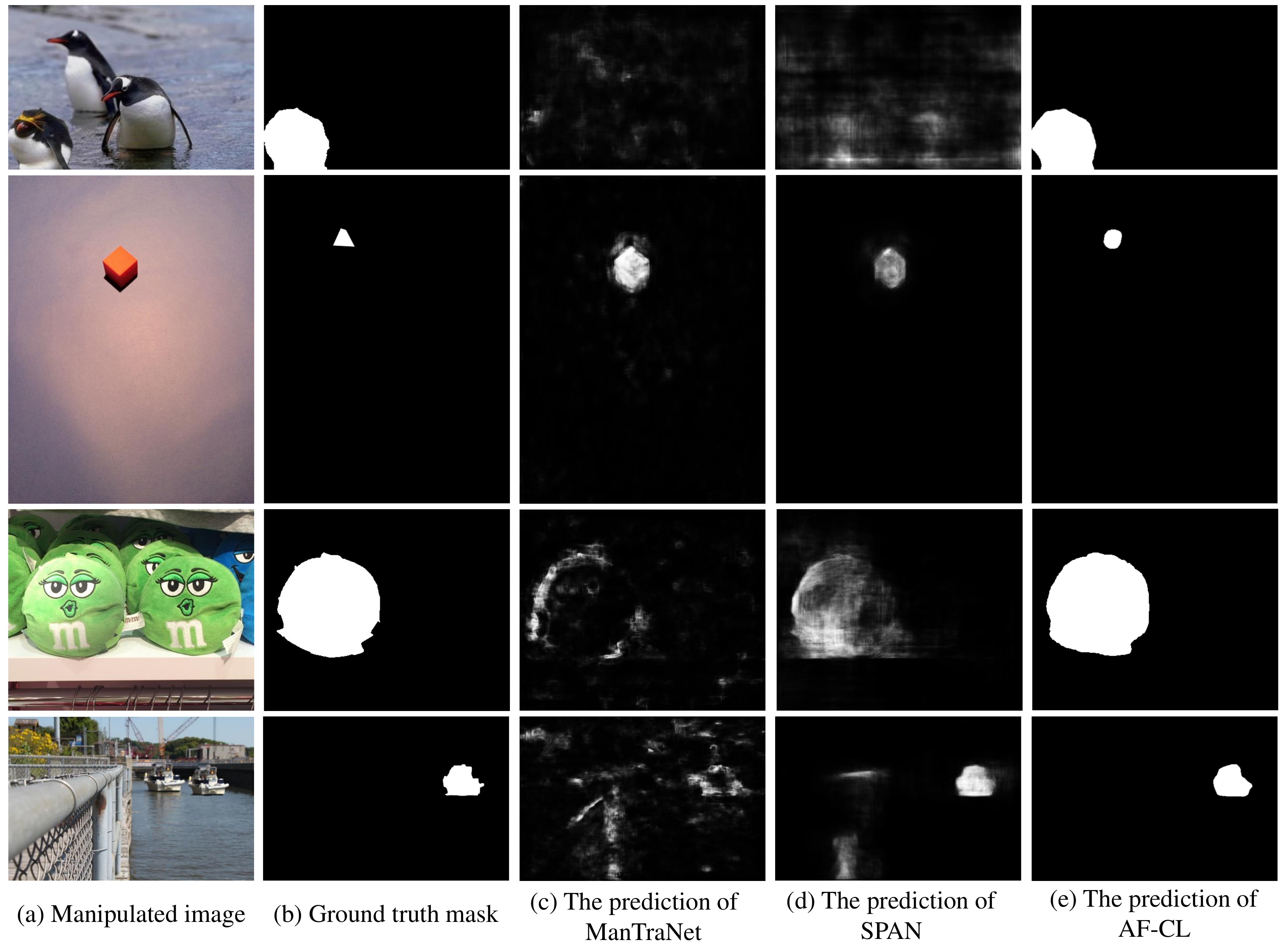}
   \caption{The visualization of predicted results of different networks on the public datasets. (a) The manipulated images. (b) Ground truth mask. (c) The prediction of ManTraNet. (d) The prediction of SPAN. (e) The prediction of AF-CL.}
   \label{Fig. 7}
\end{figure*}
In this section, we evaluate the robustness of the proposed AF-CL under different distortions, including resizing the image with different scales, gaussian blurring with different kernel size, gaussian noise with different standard deviation $\sigma$, and JPEG compression with different quality factor $q$. Note that the source code of ObjectFormer \cite{wang2022objectformer} is unavailable, and they evaluate the robustness of their pre-trained model. Therefore, we compare the effect of the distortion on the performance in NIST16 since we do not use the pre-trained model. As shown in Table \ref{Table 4}, the proposed AF-CL gains better robustness to different distortions than the ObjectFormer.

\subsection{Visualization Results}

In this section, we provide the qualitative results in Figure \ref{Fig. 7} to visually compare AF-CL with MantraNet \cite{wu2019mantra} and SPAN \cite{hu2020span}. As shown in Figure \ref{Fig. 7}, the proposed AF-CL consistently produces most accurate pixel-level localization results for manipulated region detection.

\section{Limitation}
\begin{figure}[t]
  \centering
   \includegraphics[width=0.85\linewidth]{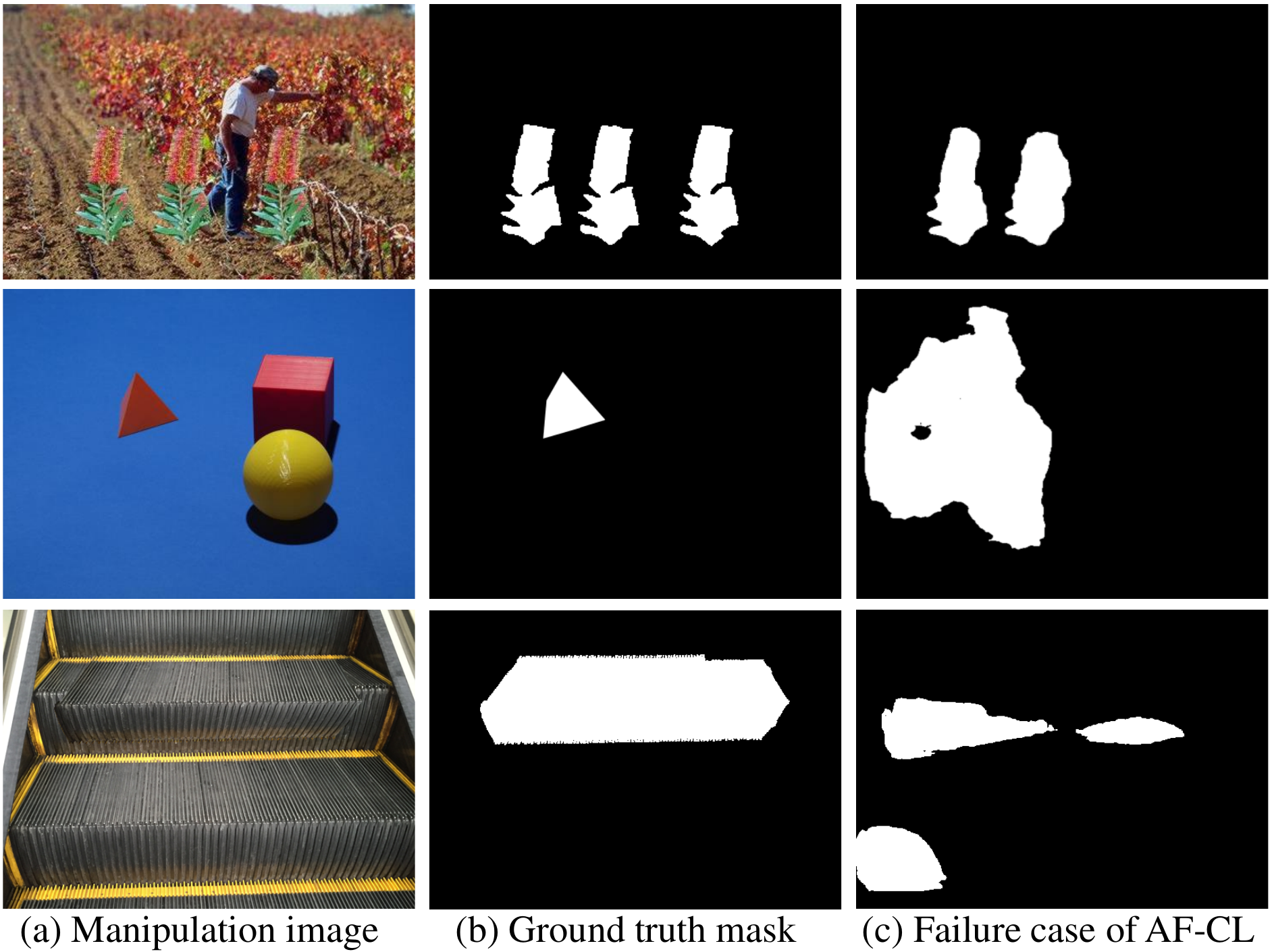}
   \caption{The failure case of AF-CL in different dataset. (a) Original manipulated images. (b) Ground truth mask. (c) The failure case of AF-CL.}
   \label{Fig. 8}
\end{figure}

Despite the proposed AF-CL showing remarkable improvements in image manipulation detection, it still has the following limitations. First, it fails to locate multi-manipulated regions (as shown in Figure \ref{Fig. 8}, first row) or all manipulated regions (as shown in Figure \ref{Fig. 8}, third row) in some cases. Second, the AUC of AF-CL is lower than state-of-the-arts on the NIST dataset. One possible reason is that some manipulation traces are complex and hard to be captured in the RGB domain.

\section{Conclusion}

In this paper, we start our research from the observations that the current manipulation detection models are simply built on the manipulation trace across the entire image, thus potentially misguiding the model to focus on irrelevant regions. Moreover, they ignore the trace relations among the pixels of each manipulation region and its surroundings. To address the above issues, we have presented a supervised contrastive learning network, called AF-CL, for image manipulation detection. The network jointly exploits MSVG and TRM to guide the model to focus on the manipulated region and its surroundings and sufficiently explore the trace relations among their pixels. The extensive experiments demonstrate that AF-CL outperforms the state-of-arts in image manipulation detection with remarkable performance improvement.

{\small
\bibliographystyle{ieee_fullname}
\bibliography{PaperForReview}
}

\end{document}